\documentclass[preprint,journal]{vgtc}       

\ifpdf
  \pdfoutput=1\relax                   
  \usepackage{graphicx}                
  \DeclareGraphicsExtensions{.pdf,.png,.jpg,.jpeg} 
\else
  \usepackage{graphicx}                
  \DeclareGraphicsExtensions{.eps}     
\fi%

\graphicspath{{figures/}{pictures/}{images/}{./}} 

\usepackage{microtype}                 
\PassOptionsToPackage{warn}{textcomp}  
\usepackage{textcomp}                  
\usepackage{mathptmx}                  
\usepackage{times}                     
\usepackage{cite}                      
\usepackage{tabu}                      
\usepackage{booktabs}                  

\usepackage{enumitem}
\usepackage{comment}
\usepackage{subfiles}
\usepackage{kotex}

\usepackage{framed,enumitem}

\newcommand{\parahead}[1]{{\vspace{0.06cm}\fontfamily{phv}\selectfont{#1} \ \ }}

\usepackage{algorithm}
\usepackage[noend]{algpseudocode}
\usepackage{subfiles}
\usepackage{kotex}
\usepackage{xspace}
\usepackage{enumerate}
\usepackage{listings}
\usepackage{amsmath}
\usepackage{gensymb}
\usepackage{tabularx}
\usepackage{makecell}
\usepackage{multirow}

\usepackage{dblfloatfix}

\def\St{Steadiness\xspace}
\def\Co{Cohesiveness\xspace}
\def\snc{Steadiness and Cohesiveness\xspace}
\def\Mg{Missing Groups\xspace}
\def\Fg{False Groups\xspace}

\usepackage{comment}

\onlineid{1020}

\vgtccategory{Research}
\vgtcpapertype{Data Transformations\vspace{-10pt}}

\title{Measuring and Explaining the Inter-Cluster Reliability \\ of Multidimensional Projections}

\author{Hyeon Jeon, Hyung-Kwon Ko, Jaemin Jo, Youngtaek Kim, and Jinwook Seo}
\authorfooter{
\item
 Hyeon Jeon, Hyung-Kwon Ko, Youngtaek Kim, Jinwook seo are with Seoul National University. E-mail: \{hj, hkko, ytaek.kim\}@hcil.snu.ac.kr, jseo@snu.ac.kr.
\item
 Jaemin Jo is with Sungkyunkwan University. E-mail: jmjo@skku.edu.
}

\shortauthortitle{Jeon \MakeLowercase{\textit{et al.}}: Measuring and Explaining the Inter-Cluster Reliability of Multidimensional Projections}

\abstract{
We propose \snc, two novel metrics to measure the inter-cluster reliability of multidimensional projection (MDP), specifically how well the inter-cluster structures are preserved between the original high-dimensional space and the low-dimensional projection space. 
Measuring inter-cluster reliability is crucial as it directly affects how well inter-cluster tasks (e.g., identifying cluster relationships in the original space from a projected view) can be conducted; 
however, despite the importance of inter-cluster tasks, we found that previous metrics, such as Trustworthiness and Continuity, fail to measure inter-cluster reliability. 
Our metrics consider two aspects of the inter-cluster reliability: \St measures the extent to which clusters in the projected space form clusters in the original space, and \Co measures the opposite. They extract random clusters with arbitrary shapes and positions in one space and evaluate how much the clusters are stretched or dispersed in the other space. 
Furthermore, our metrics can quantify pointwise distortions, allowing for the visualization of inter-cluster reliability in a projection, which we call a reliability map.
Through quantitative experiments, we verify that our metrics precisely capture the distortions that harm inter-cluster reliability while previous metrics have difficulty capturing the distortions. 
A case study also demonstrates that our metrics and the reliability map 1) support users in selecting the proper projection techniques or hyperparameters and 2) prevent misinterpretation while performing inter-cluster tasks, thus allow an adequate identification of inter-cluster structure.

} 


\CCScatlist{ 
 \CCScat{K.6.1}{Management of Computing and Information Systems}%
{Project and People Management}{Life Cycle};
 \CCScat{K.7.m}{The Computing Profession}{Miscellaneous}{Ethics}
}

\vgtcinsertpkg

\begin{document}


\firstsection{Introduction}


\maketitle

\newcommand{\revise}[1]{#1}

Can we truly trust the clusters revealed by multidimensional projections (MDP)?
One way to understand high-dimensional data in various domains is to reduce its dimensionality by MDP and thoroughly check the projection in a lower-dimensional space. 
However, distortions inherently occur when reducing dimensionality. 
Such distortions can make meaningful patterns in projections less trustworthy and can disturb users' accurate comprehension of the original data, leading to interpretation bias \cite{nonato2018multidimensional}.
Therefore, it is important to measure the overall distortions using quantitative metrics \cite{ghosh2020interpretation, nonato2018multidimensional}, or to visualize where and how the distortions occurred in the projection \cite{lespinats2011checkviz, nonato2018multidimensional}, which are as important as generating a good projection in the MDP analysis. 

This work was motivated by the following issue:
although inter-cluster tasks, which investigate the inter-cluster structures of a \revise{given dataset (i.e., how clusters are located and related) through its 2D projection}, have been regarded as the core tasks \cite{sedlmair2013empirical, martins2014visual} for using MDP, only a few previous metrics have tried to explain the distortions of inter-cluster structure represented in MDP thus far.
Through a literature review (\autoref{sec:3}), we organized three types of inter-cluster tasks: 1) identifying clusters, 2) seeking the relationship between clusters, and 3) comparing clusters of the original data based on the projected representation.
When performing these inter-cluster tasks, users must be aware of the preservation of the inter-cluster structure in the MDP.
We refer to this as the \textit{inter-cluster reliability}: how well the inter-cluster structures are preserved in the MDP. 
Several previous approaches have tried to reduce the loss of inter-cluster reliability during projection \cite{fu2019atsne, novakova2009radviz, novakova2011visualization, narayan2020density, Moor19Topological} or to explain it through visualizations \cite{cutura2020comparing, metsalu2015clustvis, liu2019latent}.
\revise{However, measuring reliability is a challenging problem, as real-world datasets' inter-cluster structures often have no ground truth, and thus the metrics that quantify ground truth reliability cannot exist.
Still, this research introduces new metrics to measure the loss of inter-cluster reliability by quantifying the preservation of randomly extracted clusters and validated their effectiveness through experiments.}

Previous local quality metrics for MDP focused on measuring point-point distortion \cite{venna2006local, chen2009local} or cluster-point distortion \cite{motta2015graph}, or quantified the preservation of predefined clusters (e.g., labels of points)\cite{joia2011local, fadel2015loch}. However, all three have difficulties measuring inter-cluster reliability. 
Point-point distortion describes a distortion that occurs in the distance between points.
Therefore, point-point distortion metrics take an intra-cluster distortion into account as the points within the cluster are close together but largely dismiss inter-cluster distortion as the points from different clusters are far away.  
Cluster-point distortion metrics instead 
describe a distortion that occurs between a cluster and its nearby points in both the original and projected spaces, 
and thus cannot consider complex inter-cluster structures that consist of multiple clusters, as they consider each cluster individually. Quantifying the preservation of the predefined clusters is also not an adequate measure for inter-cluster reliability, as many real-world datasets have no ground truth clusters.

In this paper, we propose \textit{\St} and \textit{\Co}, two metrics that quantitatively evaluate inter-cluster
reliability.
\St measures the inter-cluster reliability in the projected space (i.e., to what degree the cluster in the projection is in a steady state that reflects the actual cluster in the original space).
\Co measures the inter-cluster reliability in the original space (i.e., to what degree real clusters in the original space stand together cohesively in the projection). We first formulated three design considerations of \snc so that the metrics can adequately evaluate inter-cluster reliability by quantifying how well inter-cluster tasks can be performed accurately in MDPs (i.e., measure the potential accuracy of the inter-cluster tasks). 
To make our metrics fulfill the considerations, we defined new inter-cluster distortion types---\Fg and \Mg---and designed \snc to measure each, respectively. 
Our metrics measure inter-cluster reliability by repeatedly extracting random clusters from one space and quantifying how much the clusters have been stretched or dispersed in the opposite space. 

We also developed a reliability map to visualize inter-cluster reliability quantified by \snc within projections. 
By aggregating the underlying distortion information of each data point summarized in our metrics, the reliability map shows where and how the inter-cluster distortion occurred.

Through quantitative and qualitative experiments, we confirmed the effectiveness and usefulness of our metrics.
The quantitative experiments verified that our metrics accurately capture inter-cluster distortions and thus properly measure inter-cluster reliability, while baseline local metrics, such as Trustworthiness \& Continuity, failed to identify even apparent distortions. 
Furthermore, a qualitative case study showed that our metrics and the reliability map support a precise identification of the inter-cluster structure of the original space by helping users choose the proper projections and perform the inter-cluster tasks while being aware
of misinterpretations.

\begin{figure*}[t!]
    \centering
    \includegraphics[width=\linewidth]{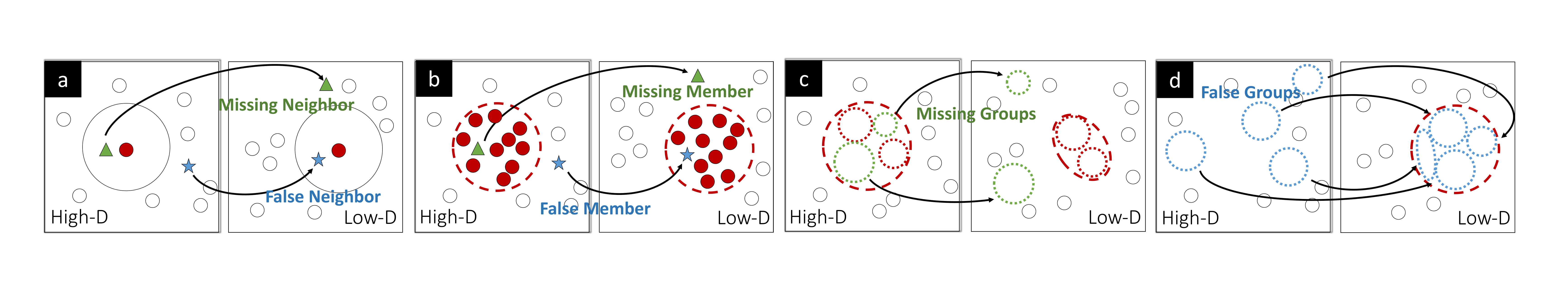}
    \caption{Illustration of the concepts of a) Missing Neighbors and False Neighbors, b) a group and its Missing / False Members, c) \Mg, and d) \Fg. Data points are represented as black-bordered circles, triangles, and stars. Groups (clusters) are depicted as the objects with dashed or dotted outlines.}
    \label{fig:distortions}
    \vspace{-6mm}
\end{figure*}

\section{Background and Related Work}

\subsection{MDP Distortions}

Many MDP techniques, such as $t$-SNE~\cite{maaten2008visualizing}, Isomap~\cite{tenenbaum2000global}, and UMAP~\cite{mcinnes2018umap}, have been proposed to understand and visualize high-dimensional data\footnote{This paper denotes both linear and nonlinear embedding of multidimensional data as MDP, following previous research \cite{nonato2018multidimensional, etemadpour2014perception, etemadpour2015user}.\vspace{-9pt}};
however, every MDP produces distortions because information loss is inevitable when dimensionality is reduced.

\subsubsection{Distortion Types}

In his seminal work~\cite{aupetit2007visualizing}, Micha{\"e}l Aupetit defined two types of MDP distortion: stretching and compression.
Stretching occurs when pairwise distances in the projected space are expanded compared to the original pairwise distances, and compression does the opposite.
Afterward, Missing Neighbors and False Neighbors \cite{lespinats2007dd, lespinats2011checkviz} distortion types were introduced to interpret the stretching and compression in the context of neighborhood preservation. 
Let $f: X \rightarrow Y$ be a smooth mapping where $X = \{ x_i \in \mathcal{R}^D, i=1,2,\ldots,N \}$ and $Y = \{ y_i \in \mathcal{R}^d, i=1,2,\ldots,N \}$ for some $D > d$.
Each data point $p_i$ has its high-dimensional coordinate, $x_i$, and the corresponding low-dimensional coordinate $y_i$.
For any point $p_i$, its $k$ neighbors in the projected and original spaces are denoted as $C_{y_{i}}$ and $C_{x_{i}}$, respectively.
Missing Neighbors are then defined as $C_{x_{i}} \backslash C_{y_{i}} $.
Similarly, False Neighbors are defined as $C_{y_{i}} \backslash C_{x_{i}} $ (\autoref{fig:distortions}a);
however, our literature review (\autoref{sec:3}) \revise{indicated} that measuring Missing and False Neighbors distortion cannot reflect how well inter-cluster tasks can be performed, and thus cannot correctly evaluate inter-cluster reliability.

To alleviate the mismatch between the inter-cluster reliability and point-point distortions, Martins et al. \cite{martins2014visual} defined distortion types relevant to the cluster-point relationship: Missing Members and False Members with regard to a group of data points. For a group $\Gamma_X \subset X$ of similar points (e.g., within the same category of a dataset or clustered by a clustering algorithm) in the original space, $\Gamma_Y \subset Y$ is used to denote the ``projected group'' that corresponds to $\Gamma_X$. Here, False Members are the points in $\Gamma_Y  \backslash  \Gamma_X$, and Missing Members are in $\Gamma_X  \backslash  \Gamma_Y$ (\autoref{fig:distortions}b); however, the literature review also \revise{revealed} that the generalization was insufficient to reflect the degree to which users can perform inter-cluster tasks precisely. We further generalize the distortion types by proposing new inter-cluster distortion types that directly harm inter-cluster reliability.

\subsubsection{Distortion Metrics}
\label{sec:22}

According to a survey conducted by Nonato and Aupetit \cite{nonato2018multidimensional}, most distortion metrics aim to measure point-point distortion. Among them, a few metrics evaluate how much Missing and False Neighbors distortion has occurred.
For instance, Trustworthiness and Continuity (T\&C)~\cite{venna2006local} locally measure how Missing and False Neighbors distort the ranks of each point's neighbors.
Mean Relative Rank Errors (MRREs)~\cite{lee2007nonlinear} are similar to T\&C; however, they consider not only the rank variance of the Missing and False Neighbors but also of True Neighbors---the points that are judged as neighbors in both spaces. 
Local Continuity Meta-Criteria (LCMC)~\cite{chen2009local} is another variant of T\&C; it only considers True Neighbors. 
Still, measuring point-point distortion cannot adequately measure inter-cluster reliability, since it needs to quantify the relationship between clusters.

Motta et al. \cite{motta2015graph} proposed graph-based group validation, which is the only metric measuring cluster-point distortion we could find as a relevant work. 
The metric first extracts clusters from both the original and projected spaces using graph-based clustering. The metric then calculates each cluster's structural persistence in the opposite space by measuring how much Missing and False Members distorted the cluster.
Given that the metric examines each cluster independently, it is inappropriate to use it to measure inter-cluster reliability, which refers to multiple clusters at once.

Measuring the distortion of predefined clusters with a clustering quality metric has been widely used to evaluate MDP. For example, Joia et al. \cite{joia2011local} and Fadel et al. \cite{fadel2015loch} used the silhouette coefficient \cite{rousseeuw1987silhouettes} to quantify cluster preservation in MDP. However, one limitation is that the inter-cluster structures of real-world datasets are usually unknown. 
Graph-based group validation also suffers from the same problem, as it performs clustering once for all data and uses it as predefined clusters.
By contrast, our metrics consider
the complex inter-cluster structure by examining repeatedly extracted random clusters, thus much accurately quantify inter-cluster reliability.

\subsubsection{Distortion Visualizations}

To overcome the inherent limitation of metrics that describe only the overall distortions with one or two representative numerical values, complementary visualizations are proposed \cite{nonato2018multidimensional}. The visualizations aim to reveal the submerged distortion information summarized by a single or two values, thus helping users identify trustworthy areas of the projection or detect distortion patterns. 

Distortion visualizations commonly highlight regions with local point-point distortions by decomposing the area into grids, where each grid cell corresponds to the data point and encodes the corresponding point's distortion to the cells.
The decomposition is usually done using a heatmap \cite{seifert2010stress}, Voronoi diagram \cite{aupetit2007visualizing, lespinats2011checkviz, heulot2012proxiviz}, or 2D point-cloud \cite{martins2014visual, martins2015explaining}. By contrast, MING \cite{colange2019interpreting} explains False and Missing Neighbors by visualizing the shared amount of the nearest neighbor graphs constructed in the original and projected space. 
In this work, we quantified pointwise distortions by aggregating the inter-cluster distortion of the clusters and visualized them.

\subsection{Inter-Cluster Reliability}

As many MDP techniques intentionally focus on local neighbors, they have trouble reflecting the original high-dimensional space's global inter-cluster structure. 
For example, Barnes-Hut $t$-SNE \cite{van2014accelerating} and LargeVis \cite{tang2016visualizing} 
concentrate on local structures by interpreting data based on $k$-Nearest Neighbor ($k$NN) graphs.
Using a $k$NN with a small $k$ ($k \ll N$) also allows them to reduce computation. However, 
as $k$NN graphs with small $k$ only maintains the relations between the point and its local neighbors, they can only reflect limited local structures \cite{fu2019atsne}.

Recently proposed MDP techniques have tried to preserve both the local and global inter-cluster structures.
For example, 
Narayan et al. \cite{narayan2020density} introduced den-SNE and densMAP, which modify $t$-SNE and UMAP, respectively, to better preserve clusters' density. 
\revise{Another common strategy is to first construct a global skeletal layout using representative points (i.e., landmarks) and formulate local structure around each landmark~\cite{fu2019atsne, joia2011local, paulovich2010two, fadel2015loch, pezzotti2016hierarchical}}.
However, even for these approaches, completely retaining the original space's inter-cluster structure during the projection is inherently impossible. Therefore, it is vital to measure the extent to which these techniques preserve inter-cluster reliability for a proper evaluation and analysis. 

Previous studies have attempted to explain the inter-cluster reliability of MDP through visualizations. For instance, the Compadre system \cite{cutura2020comparing} enables an inter-cluster structure analysis based on matrix visualization, and ClustVis \cite{metsalu2015clustvis} does so with a heatmap. Visual analytics systems \cite{chatzimparmpas2020t, liu2019latent} with similar goals have also been proposed.
Unlike these previous works, which utilized separate visual idioms to show the distortion, we adopted a strategy of visualizing the distortion within the projection \cite{lespinats2011checkviz, martins2014visual} to explain inter-cluster reliability. 
Therefore, users can directly identify where and how the inter-cluster distortions occurred in the projection.

\section{Design Considerations for \snc}

In this section, we first survey inter-cluster tasks, which are essential in data analysis using MDP \cite{martins2014visual, sedlmair2013empirical}, through a literature review.
We then establish the design considerations based on the survey that our metrics (\St and \Co) should satisfy to adequately measure how much inter-cluster tasks can be held precisely in MDP.

\subsection{Inter-Cluster Tasks Analysis}
\label{sec:3}

To identify the importance of inter-cluster structure preservation and to elicit the design considerations for our metrics, we inspected previous papers that addressed tasks related to clusters. We first investigated 31 papers introduced in a systematic review conducted by Sacha et al. \cite{sacha2016visual}, which surveyed how analysts interact with MDP.
We also investigated 155 articles citing Sacha et al. using Google Scholar to expand the search space.
As a result, we identified 26 papers concerning inter-cluster tasks: \revise{tasks that investigate the inter-cluster structure of original data through its 2D projections}. 
Regarding the task taxonomy for MDP proposed by Etemadpour et al. \cite{etemadpour2014perception, etemadpour2015user}, we classified the tasks into three categories in terms of inter-cluster distortions.
We then organized them into individual tasks, as listed in the following:

\setlist{topsep=0.1em}
\begin{itemize}[noitemsep] \setlength
    \item[\textbf{T1:}]\textbf{Identify separate clusters in the original space by exploring clusters in the  projected space}.
    Recognize the separation between clusters \cite{choo2010ivisclassifier, endert2011observation, wang2017perception} or distinguish a cluster from the others \cite{poco2011framework, nam2007clustersculptor}. 
    \item[\textbf{T2:}]\textbf{Seek the relationships between clusters of the original space based on those of the projected space}.
    \textbf{(1)} Investigate the hierarchical or inclusion relation between clusters
    \revise{(i.e., check whether clusters can again be divided into  smaller parts with higher density, which we call ``subclusters'')}
    \cite{liu2014distortion, xia2017ldsscanner}. \textbf{(2)} Estimate the clusters' similarities based on their distances in the projected space \cite{nam2007clustersculptor, wenskovitch2020respect}.
    \item[\textbf{T3:}]\textbf{Compare clusters in the original space based on their features in the projected space}.
    Estimate and compare the clusters' original sizes or densities based on their sizes or densities in the projected space \cite{chatzimparmpas2020t, amabili2017visualizing}.
\end{itemize}

The tasks were verified through semi-structured interviews with four machine learning (ML) engineers (E1-E4) with more than three years of experience. Three engineers confirmed that they practically perform the tasks for the real-world problem. Only E1 said that he does not perform the tasks. This is because he usually works with data with well-distributed vector representations processed by a deep neural network, where no inter-cluster structure exists.

Previous surveys of high-dimensional data analysis tasks based on MDPs further confirm our task analysis results, as those works show  similar results to ours, despite using different methodologies. T1 is covered by Brehmer et al.'s task taxonomy based on interviews with 10 data analysts \cite{brehmer2014visualizing}, and T2 and T3 are covered by the taxonomy of cluster separations in MDPs discussed by Sedlmair et al. \cite{sedlmair2012taxonomy}. 

Our survey \revise{indicated} that point-point and cluster-point distortion metrics cannot correctly quantify how well inter-cluster tasks can be performed. 
Point-point distortion metrics focus on each point's neighborhood instead of the inter-cluster structure. Therefore, the metrics can only measure the potential accuracy of relation-seeking tasks relevant to point-point relations, such as finding $k$NN of the given point \cite{etemadpour2015user}; they cannot measure the extent to which inter-cluster tasks can be performed accurately as those tasks focus on the cluster level. 

Cluster-point distortion metrics can estimate the potential accuracy of T3, as the size and density of each cluster are related to the cluster itself. 
More precisely, if an MDP generates outliers for a cluster, the cluster's size is reduced (if the density is maintained), or its density is reduced (if the size is maintained). Both distortions directly affect the comparison task.
By contrast, cluster-point distortion metrics still fails for T1 and T2. 
As the metrics consider each cluster independently, they can only work for cluster identification tasks related to a single cluster (e.g., distinguishing the outliers of the cluster \cite{etemadpour2015user}) or relation-seeking tasks about a single cluster (e.g., finding $k$ closest points of the given cluster \cite{etemadpour2015user}); however, they cannot provide sufficient information required 
to support T1 and T2 that consider multiple clusters at once.

\subsection{Design Considerations}

\label{sec:design}

Based on the task analysis, we formulated three design considerations (C1, C2, C3) that \snc should satisfy to adequately quantify how accurately the three inter-cluster tasks can be performed  and thus able to precisely measure inter-cluster reliability. 

\begin{itemize}[noitemsep] \setlength\itemsep{0em}
    \item [\textbf{C1:}]\textbf{Capture the inter-cluster structure in detail}.
    The inter-cluster structure in MDP is complex and intertwined \cite{xia2017ldsscanner}, and often has no ground truth. Furthermore, each cluster's characteristics (e.g., shape, density, or size) vary widely \cite{harel2001clustering}. 
    Therefore, to quantify how precisely users can identify clusters (T1) or seek relationships between them (T2), we should thoroughly consider the inter-cluster structure in detail.
    \item [\textbf{C2:}]\textbf{Consider stretching and compression individually}.
    The distances between clusters may be affected by two aspects of geometric distortions: stretching and compression \cite{aupetit2007visualizing}. 
    If stretching occurs, users can misunderstand nearby clusters as distinct clusters. The opposite can happen if compression occurs (i.e., nearby groups can be identified as a single cluster). 
    Furthermore, clusters' size and density can be overestimated due to stretching or can be underestimated by compression.
    As the two aspects of distortion result in different types of misperceptions about the clusters' size and density (T3) or their distance (T2-2), we should consider both aspects individually.
    \item [\textbf{C3:}]\textbf{Measure how accurately the clusters identified in the projection reflect their original density and size}.
    Users can have misconceptions when comparing clusters (T3) if the projected clusters' size and density do not reflect those in the original space. To correctly quantify how much such misunderstandings can happen, we need to measure how accurately the clusters in the projection reflect their original density and size.
\end{itemize}

\section{\snc}
\label{sec:4}

We propose \snc to measure inter-cluster reliability by evaluating inter-cluster distortion, satisfying our four design considerations.
\St measures inter-cluster reliability in the projected space (e.g., separated clusters in the original high-dimensional space are still separated in the projected space), while \Co does the same for the original space (e.g., each cluster in the original space is not dispersed in the projected space).

\subsection{Defining Inter-Cluster Distortion Types}

To design \snc, we first defined two inter-cluster distortions types---\Fg and \Mg---by generalizing False and Missing Neighbors to the cluster level.
False Groups distortion denotes the cases in which a low-dimensional group in a single cluster (red dashed circle in \autoref{fig:distortions}d) consists of separated groups in the original space (blue dotted circles in \autoref{fig:distortions}d), and Missing Groups distortion occurs when the original group (red dashed circle in \autoref{fig:distortions}c) misses its subgroups (green dotted circles in \autoref{fig:distortions}c) and therefore is divided into multiple separated subgroups in the projected space. 
\snc evaluate how well projections avoid False and Missing Groups, respectively (C2).

\subsection{Computing \snc}
\label{sec:4snc}

We compute inter-cluster reliability through the following procedure: 
(Step 1) Constructing dissimilarity matrices. (Step 2) Iteratively computing partial distortions. (Step 3) Aggregating partial distortions into \snc.
Based on the definitions of the two measures,
\St increases as clusters extracted from the projected space stay closer consistently together in the original space. In contrast, \Co increases when clusters in the original space are maintained more consistently in the projected space. 

Each step is designed to satisfy all the design considerations (\autoref{sec:design}). First, we split the workflow to handle \snc independently after step 1 (C2). Step 2 exploits randomness to cover the complex inter-cluster structures (C1) and inherently quantifies how well the original density and size of clusters are retained (C3).

The workflow requires four functions as hyperparameters: 
\vspace{-2.3mm}
\setlist{}
\begin{itemize}[noitemsep]
    \setlength{\itemsep}{0pt}
    \item Distance function for points, \texttt{dist}
    \begin{itemize}[noitemsep, nolistsep]
        \item Input: two points $p$ and $q$
        \item Output: the distance (or dissimilarity) between $p$ and $q$
    \end{itemize}
    \item Distance function for clusters, \texttt{dist\_cluster}
    \begin{itemize}[noitemsep, nolistsep]
        \item Input: two clusters $A$ and $B$
        \item Output: the distance (or dissimilarity) between $A$ and $B$
    \end{itemize}
    \item Cluster extraction function, \texttt{extract\_cluster}
    \begin{itemize}[noitemsep, nolistsep]
        \item Input: a seed point $p$
        \item Output: a cluster in the projected space (for \St) or the original space (for \Co) centered on $p$
    \end{itemize}
    \item Clustering function, \texttt{clustering}\
    \begin{itemize}[noitemsep, nolistsep]
        \setlength{\parskip}{-1pt}
        \item Input: a set of points $P$
        \item Output: a clustering result of the input points $P$ where the clustering takes place in the original space (for \St) or the projected space (for \Co)
    \end{itemize}
\end{itemize}
\vspace{-2.5mm}
Two distance functions are used to compute the amount of inconsistency, while the other two functions are used for the iterative computation of partial distortions. 
These functions are explained in detail in \autoref{sec:4hyper}.


\subsubsection{Step 1: Constructing Dissimilarity Matrices}

We begin the measurement by constructing dissimilarity matrices $D^+$ and $D^-$. 
We first construct distance matrices $H$ and $L$ satisfying $H_{ij} = \texttt{dist}(h_i, h_j)$ and $L_{ij} = \texttt{dist}(l_i, l_j)$,
where $h_i$ and $l_i$ denote the original and projected coordinates of input data point $p_i$, respectively. 
For \texttt{dist}, we used Shared-Nearest Neighbor (SNN) based dissimilarity \cite{ertoz2003finding} as a default (\autoref{sec:4hyper}).
$H$ and $L$ are then normalized by dividing all elements by their max elements $H_{max}$ and $L_{max}$. 
Raw dissimilarity matrix $D$ is obtained by subtracting $L$ from $H$. The positive elements in $D$ denote that the distance between the corresponding points pair is compressed, and the opposite denotes that the distance is stretched. We then construct $D^+$ and $D^-$, where $D^+_{ij} = (D_{ij}$ if $D_{ij} > 0$, else $0)$ and $D^-_{ij} = (- D_{ij}$ if $D_{ij} < 0$, else $0)$.

\subsubsection{Step 2: Iteratively Computing Partial Distortions}

\label{sec:422}

The next step is to iteratively compute partial distortions by randomly extracting clusters from one space and evaluating their dispersion in the opposite space. 
In this section, we describe how to compute partial distortions in a single iteration.

\parahead{Extracting random clusters}
For each iteration, we first select a seed point randomly in the projected space (\St) or the original space (\Co).
Then, the \texttt{extract\_cluster} function takes the random seed point as input and extracts a cluster centered on the point as output. 
The random selection of the seed point leads to the extraction of clusters from diverse locations and therefore it is possible to cover the entire inter-cluster structure after sufficient iterations (e.g., 200 iterations for the data consists of 10,000 points) (C1). By default, we use the SNN similarity (\autoref{sec:4hyper}) for the \texttt{extract\_cluster} function to gather points near the seed point. 

\parahead{Revealing the cluster's dispersion in the opposite space}
Next, we reveal how the randomly extracted cluster is dispersed in the opposite space.
To do this, the \texttt{clustering} function takes the points of the extracted cluster generated by \texttt{extract\_cluster} as input. Afterward, the \texttt{clustering} function clusters the input points in the opposite space and returns the set of separated clusters $\mathcal{C} = \{ C_1, C_2, \ldots, C_n \}$ as output. 
Hierarchical DBSCAN (HDBSCAN) \cite{campello2013density, mcinnes2017hdbscan} utilizing an SNN-based distance function is used as the default \texttt{clustering} function (\autoref{sec:4hyper}).

This step also allows the metrics to measure how well the clusters reflect their original density and size (C3). 
If a cluster's original outliers are merged into a single cluster during MDP (\Fg distortion), either the cluster's size or density will be increased. This situation can be captured while checking the projected cluster's dispersion in the original space. 
For the opposite case (\Co), if an original space's cluster loses some of its points during MDP, either its size or density in the projected space will be reduced. Revealing \Mg distortion captures this issue (\autoref{sec:71}). 

\parahead{Computing distortions between dispersed groups}
In this step, we take $\mathcal{C}$ as input and generate distortion $m_{ij}$ and its weight $w_{ij}$ for each pair of clusters ($C_i, C_j$), based on point-stretching and point-compression metrics proposed by Micha{\"e}l Aupetit \cite{aupetit2007visualizing}. 
We generalized the point-stretching and point-compression to the cluster-stretching $m_{ij}^{stretch}$ (\St) and cluster-compression $m_{ij}^{compress}$ (\Co) by substituting the distance between points to the distance between clusters. 
For each cluster pair ($C_i$, $C_j$), we compute their distance $\delta_h$ and $\delta_l$ in the projected space and the original space, respectively, utilizing \texttt{dist\_cluster}. The default \texttt{dist\_cluster} is designed by expanding the SNN-based distance function for points (\autoref{sec:4hyper}).
Then, we check whether the distance is compressed or stretched and consecutively compute the distortion as follows:
\vspace{-5pt}
\[
m_{ij}^{\text{stretch}} = \frac{\mu_{C_i, C_j}^{stretch} - \min{D^-}}{\max{D^-} - \min{D^-}}, \text{ }
m_{ij}^{\text{compress}} = \frac{\mu_{C_i, C_j}^{compress} - \min{D^+}}{\max{D^+} - \min{D^+}},
\vspace{-5pt}
\]
where 
\[
\vspace{-5pt}
\mu_{C_i, C_j}^{stretch} = - (\delta_h - \delta_l) \text{ if } - (\delta_h - \delta_l) > 0 \text{, otherwise } 0, 
\]
\vspace{-5pt}
\[
\mu_{C_i, C_j}^{compress} = \delta_h - \delta_l \text{ if } \delta_h - \delta_l > 0 \text{, otherwise } 0.
\]
\vspace{-5pt}

The weight $w_{ij}$ of a pair $(C_i, C_j)$ is determined as $|C_i| \cdot |C_j|$. 
The weights penalize the distortion of larger clusters more than smaller ones; thus, we can deal with the inter-cluster structure consisting of the clusters of various sizes (C1).

\subsubsection{Step 3: Aggregating Partial Distortions}

This step aggregates the iteratively computed partial distortions to \snc. 
The iterative partial distortion measurement generates a set of distortions and their corresponding weights. Let's denote the set as follows: 
\begin{itemize}[leftmargin=*, itemsep=0.05pt]
    \item $(m^{compress}_1, w_1), \cdots , (m^{compress}_{n_{St}}, w_{n_{St})}$ where $n_{St}$ denotes the number of total cluster pairs generated throughout the entire partial distortion measurement of \St.
    \item $(m^{stretch}_1, w_1), \cdots , (m^{stretch}_{n_{Co}}, w_{n_{Co})}$ where $n_{Co}$ denotes the number of total cluster pairs generated throughout the entire partial distortion measurement of \Co.
\end{itemize}
We then calculate the final scores as follows: 
\begin{itemize}[leftmargin=*]
    \item[] $\textsc{\St} = 1 - \sum_{i=1}^{n_{St}}w_i \cdot m_i^{compress} / \sum_{i=1}^{n_{St}}w_i$.
    \item[] $\textsc{\Co} = 1 - \sum_{i=1}^{n_{Co}}w_i \cdot m_i^{stretch} / \sum_{i=1}^{n_{Co}}w_i$.
\end{itemize}
The final scores lie in the range of [0, 1]. The weighted average is subtracted from 1 to assign lower scores to lower-quality projections.

\subsection{Designing Hyperparameter Functions}

\label{sec:4hyper}

\subsubsection{Parameterizing Hyperparameter Functions}

The workflow of computing \snc requires four hyperparameter functions: \texttt{dist}, \texttt{dist\_cluster}, \texttt{clustering}, and \texttt{extract\_cluster}. 
We parameterized these functions because both the definition of distance and the definition of clusters vary depending on the analysis goals. 
There are various ways to define the distance between two data points (e.g., Euclidean distance, geodesic distance, cosine similarity). 
The definition of clusters also varies, and thus many different clustering algorithms (e.g., \textit{K}-Means \cite{duda1973pattern}, Density-based clustering \cite{ester1996density}, Mean shift\cite{comaniciu2002mean}) exist. 
Therefore, it is unreasonable to use a fixed definition for both.
This is in line with the fact that, as there are various ways to define similarity between each point and its local neighbors, there are diverse local metrics that utilize different similarity definitions. 
However, parameterization could reduce metrics' interpretability. Thus, we designed the default hyperparameter functions \revise{that align with our design considerations} to allow users to easily understand and use our metrics.

\subsubsection{Default Hyperparameter Functions}

To design the default functions, we first set the definitions of distance and cluster.
We defined distance as the dissimilarity of the points based on the Shared-Nearest Neighbors (SNN) \cite{ertoz2003finding} similarity, which assigns a high similarity to point pairs sharing more $k$NNs. 
SNN-based dissimilarity was selected because \snc should reflect the inter-cluster structure of the original high-dimensional space.
Although it is common to use a $k$NN graph to reflect a high-dimensional space \cite{van2014accelerating, tang2016visualizing}, 
\revise{$k$NN's ability to describe the structure of data decreases as dimensionality grows \cite{beyer1999nearest, hinneburg2000nearest}. }
SNN-based dissimilarity tackles this issue as the similarity of two points is robustly confirmed by their shared neighbors, thus better representing the structure of high-dimensional spaces compared to $k$NN \cite{ertoz2002new, liu2018shared}. 
We also defined a cluster as the contiguous data region, or manifold with  an arbitrary shape, where the density of the region is higher than its surroundings. This definition is followed by density-based clustering algorithms. We used this definition because the metrics should capture the complex and intertwined inter-cluster structure consisting of clusters of various sizes and shapes (C1), and therefore 
should be able to define clusters more flexibly. We designed the default hyperparameter procedures to satisfy both the definitions and the original design considerations (\autoref{sec:design}).

\parahead{Distance function for points, \large{\texttt{dist}}}
As mentioned, the distance function is based on SNN similarity. Let us first denote $k$-nearest neighbors of a point $p$ as $p_1, p_2, \cdots , p_k$, in order.
The SNN similarity between two points $p, q$ is defined as $sim(p, q) = \sum_{(m,n) \in N_{p,q}}(k + 1 - m) \cdot (k + 1 - n)$ 
where $N_{p,q}$ is a set of each pair $(m,n)$ satisfying $p_m = q_n$. 
$sim(p, q)$ increases when more $k$-nearest neighbors with high ranks overlap. 
We consecutively normalized all SNN similarity values by dividing them by the max SNN similarity \texttt{max\_sim} of the dataset. 
Finally, we defined distance function \texttt{dist} as \texttt{dist}$(p,q)  = 1 / (sim(p,q) + \alpha)$. We used reciprocal transformation \cite{tan2016introduction} to further penalize low similarity, where $\alpha$ controls the amount of penalization. $\alpha = 0.1$ is used as the default.

\parahead{Distance function for clusters, \large{\texttt{dist\_cluster}}} As for \texttt{dist\_clust-er}, we first defined the similarity between clusters and converted it to their distance. 
\revise{We used average linkage \cite{murtagh2012algorithms}, as it is robust to outliers compared to competitors (e.g., simple linkage),} thus defining the similarity of two clusters $A$ and $B$ as $sim(A, B) = \sum_{p_A \in A} \sum_{p_B \in B} sim(p_A, p_B)/|A|\cdot|B|$,
where $p_A, p_B$ denotes the points in $A$ and $B$. We then defined the distance between $A$ and $B$ as \texttt{dist\_cluster}$(A,B) = 1 / (sim(A,B) + \alpha)$.

\parahead{Clustering function, \large{\texttt{clustering}}} As our definition of cluster is the one used in conventional density-based clustering, designing \texttt{clustering} required a single decision: selecting the proper density-based clustering algorithm.
We selected HDBSCAN, which is a state-of-the-art density-based clustering algorithms. 
As HDBSCAN can handle clusters with various shapes and densities and is robust to noises (outliers) \cite{mcinnes2017accelerated}, exploiting it helps to reveal the dispersion of clusters regardless of the clusters' characteristics (e.g., shape, size, or density). Therefore, it helps the metrics deal with complex inter-cluster structures (C1). 
HDBSCAN also tackles the curse of dimensionality \cite{vijendra2011efficient}, which suits our metrics that need to consider the higher dimensional space. 
To align \texttt{clustering} with our dissimilarity definition, our HDBSCAN utilized \texttt{dist} for the distance calculation.

\parahead{Cluster extraction function, \large{\texttt{extract\_cluster}}}
The design of \texttt{extract\_cluster} mainly follows a density-based clustering process, aligned to \texttt{clustering}; it uses random seed point as a sole core point and assigns nearby points, \revise{which are treated as non-core points}, successively to form a cluster. In detail, the function traverses seed point $p$'s $k$-nearest neighbors and includes each neighbor point $p_i$ as a cluster member with a probability of $sim(p, p_i) /$\texttt{max\_sim}. When the neighbor point is determined as a cluster member, it goes into a queue so that its neighbor can also be traversed later. \revise{Adding neighbors stochastically makes extracted clusters not span the entire $k$NN graph but form a dense structure.}

To diversify the size of the extracted clusters, we limited the traversal number starting from the seed point and allowed repeated visits. Combined with the random starting seed point, this strategy enriches the range that our metrics cover, thus helping the metrics deal with complex inter-cluster structures (C1). The strategy fundamentally relies on the fact that randomness can help analyze a complex, uncertain system  \cite{tempo2012randomized}. We fixed the number of traversal to 40\% of the total number of data points for our evaluations (\autoref{sec:6quant}, \ref{sec:7777}).

\begin{figure*}[ht]
    \centering
    \includegraphics[width=\textwidth]{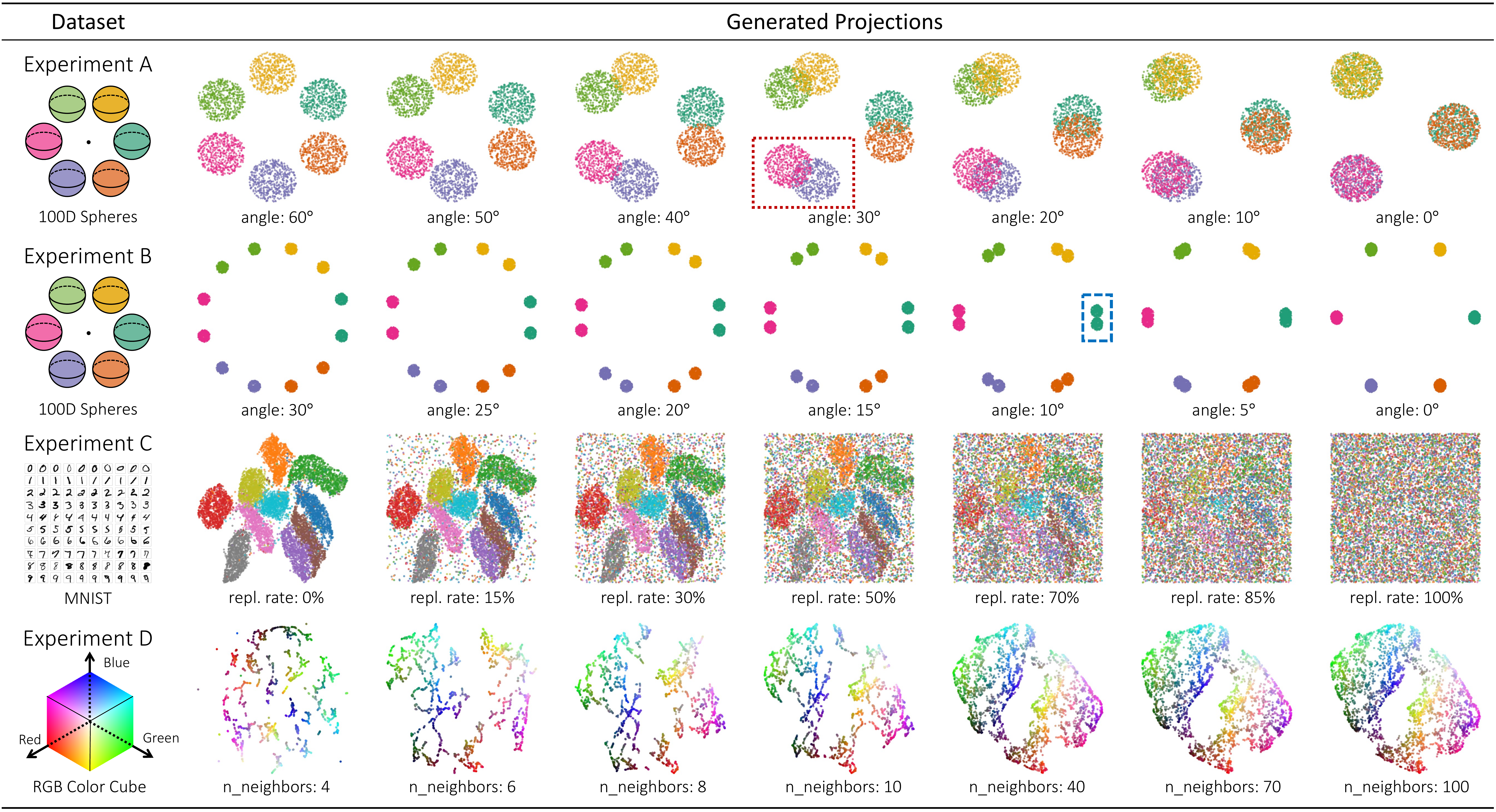}
    \caption{The datasets and their projections used in Experiment A-D. 
    A, B) The synthetic projection of the dataset with six 100-dimensional spheres, consists of six circles (A) and 12 circles (B) with an increasing amount of overlapping. 
    C) \textsc{MNIST} dataset and their projections created by replacing a certain proportion of the $t$-SNE projection with random points with increasing replacement (repl.) ratios.
    D) The UMAP projection of the data randomly sampled from the RGB color cube dataset with an increasing \texttt{nearest neighbors} hyperparameter value.
    }
    \vspace{-3mm}
    \label{fig:projections}
\end{figure*}

\subsection{Visualizing \snc}

\label{sec:5}

To overcome the limitation of metrics in that they describe the overall distortion in a single or two numeric values, we developed a complementary visualization: a reliability map (\autoref{fig:expexp}, \ref{fig:mnists}, \ref{fig:fmnists}). The reliability map reveals how and where inter-cluster distortion occurred by showing \snc at each point.
The distortions at each point are quantified by aggregating partial distortion values computed throughout the measurement of our metrics. The map shows these pointwise distortions embedded within the projection. 

The pointwise distortion is obtained by aggregating partial distortions computed throughout the iterative process (\autoref{sec:422}).
Recall that the iterative computation results in a set of distortion $m_{ij}^{compress}$ or $m_{ij}^{stretch}$ and weight $w_{ij}$ between a pair of clusters $\{C_i, C_j\}$. 
For all $\{C_i, C_j\}$, we register every $p \in C_j$ to every $q \in C_i$ with the distortion strength $m_{ij} \cdot w_{ij}$, and do the same in the opposite direction.
Duplicated registration of a point are removed by averaging distortion strengths.
We compute each point's approximated local distortion by summing up the registered distortion strengths.

The reliability map visualizes these pointwise distortions through edge-based distortion encoding. 
\revise{We constructed a $k$NN graph in the projection and made each edge $(p,q)$ of the graph}
depict the sum of $p$ and $q$'s pointwise distortion. 
If the points within a narrow region have high distortion, the edges between the points will be intertwined in the region (e.g., red dotted contours in \autoref{fig:mnists}, \ref{fig:fmnists}); they will be recognized as clusters with distinguishable inter-cluster distortion. 
\revise{However, using a large $k$ might generate visual clutter;} we empirically found that $k$ between 8 and 10 is an adequate choice for both expressing inter-cluster distortion and \revise{avoiding visual clutter.}
\revise{Martins et al.'s point cloud distortion visualization \cite{martins2014visual} is similar to ours, but it computes the distortion value at each pixel instead of encoding to edges.}

To express \Fg and \Mg distortion types simultaneously, we used CheckViz's two-dimensional color scale \cite{lespinats2011checkviz} (lower right corner of \autoref{fig:mnists}).
Following the color scheme of CheckViz, we assigned purple to the edges with \Fg distortion and green to the edges with \Mg distortion. Moreover, edges with no distortion are represented as white, while black edges indicate that both distortion types occurred together. 

We also implemented a cluster selection interaction (e.g., lower right box in \autoref{fig:expexp}) to allow users to identify \Mg distortion more precisely.
After users select a cluster $C=\{p_1, p_2, \cdots, p_{|C|}\}$ by making a lasso with a mouse interaction, the reliability map constructs $C'=\{A(p_1) \cup A(p_2) \cup \cdots \cup A(p_{|C|})\}$, where $A(p)$ denotes the set of registered points of a point $p$. 
Subsequently, the edges connected to the points in $C'$ are highlighted in red. Each highlighted edge's opacity encodes the sum of distortion strength of its incident points toward $C$ (i.e., how much distance between its incident points and $C$ is stretched).

\section{Implementation}

\snc are written in Python with an interface for users or programmers to easily implement and use user-defined hyperparameter functions. 
This is to facilitate the development and verification of possible alternatives of \snc later. The partial distortion computation is parallelized with CUDA GPGPU \cite{nickolls2008scalable} supported by Numba \cite{lam2015numba}. We implemented the reliability map in JavaScript using D3.js \cite{bostock2011d3}. 
\revise{The source code of the metrics and the map is available at \href{https://github.com/hj-n/steadiness-cohesiveness}{\texttt{github.com/hj-n/steadiness-cohesiveness}} and \href{https://github.com/hj-n/snc-reliability-map}{\texttt{github.com/hj-n/snc-reliability-map}}}, respectively.

\section{Quantitative Evaluations and Discussions}

\label{sec:6quant}

We evaluated \snc in terms of quantifying inter-cluster reliability by comparing them with existing local distortion metrics.
We verified that our metrics well capture inter-cluster reliability, while previous local metrics miss some cases even with the apparent distortions. 
The reliability map ascertained that our metrics accurately captured where and how the inter-cluster distortion occurred. 
Moreover, we evaluated our metrics' robustness by testing simpler hyperparameter functions (\autoref{sec:4hyper}). 

As baseline metrics, we chose T\&C and MRREs (\autoref{sec:22}), the two representative local metrics that measure nearest neighbors preservation. 
We chose the two for comparison because 1) they were designed to measure Missing and False Neighbors, the point-wise version of Missing and False
Groups and 2) nearest-neighbor preservation has been used as the core evaluation criteria for evaluating MDP techniques previously \cite{pezzotti2016hierarchical, van2014accelerating, fu2019atsne, lee2011shift, Moor19Topological}.
For MRREs, in this section we use ``MRRE [Missing]'' for the one that quantifies Missing Neighbors, and ``MRRE [False]'' for the other that quantifies False Neighbors.

\subsection{Sensitivity Analysis}

We conducted four experiments to check whether \snc can sensitively measure inter-cluster reliability. 
We designed the first two experiments (A, B) to evaluate our metrics' ability to quantify the inter-cluster distortion using the projections with synthetically generated \Fg (Experiment A) or \Mg (Experiment B) distortions respectively.
The next two experiments (C, D) were conducted to investigate whether our metrics have the ability to properly assess the overall inter-cluster reliability difference of the projections.

\subsubsection{Experimental Design}

\parahead{Experiment A: Identifying \Fg} The goal of the first experiment was to evaluate whether and how \St and previous local metrics (Continuity, MRRE [False]) identify \Fg.
We first generated high-dimensional data consisting of six 100-dimensional spheres whose centers were equidistant from the origin.
Each sphere consisted of 500 points.
We then set the initial 2D projection of the dataset as six  circles around the origin (the first projection on the first row of \autoref{fig:projections}). 
Note that this projection is the most faithful view of the original data as we made each circle correspond to one high-dimensional sphere.
To simulate \Fg distortion, we then distorted this ground-truth projection by overlapping the circles in pairs (the first row of \autoref{fig:projections}).
For each pair of circles $(A, B)$ centered at $C_A$, $C_B$, respectively, we adjusted the degree of overlap by changing $\angle C_A O C_B$ from $60\degree$ to $0\degree$ with an interval of $2.5\degree$, where $O$ is the origin. 
For each projection, we measured \snc ($k=[80, 90, 100, 110, 120]$,  500 iterations), T\&C ($k=[5, 10, 15, 20, 15]$), and MRREs ($k=[5, 10, 15, 20, 15]$). 
We used different $k$ values and used the mean of their results as the final score for soundness.

\parahead{Experiment B: Identifying \Mg} To evaluate \Co and previous local metrics' (Trustworthiness, MRRE [Missing]) ability to measure \Mg distortion, we used the same high-dimensional dataset as Experiment A, but this time, we synthesized the initial projection consisting of 12 equally distant circles, where each consists of 250 points. We made a pair of nearby circles correspond to a single sphere in the original space (the second row of \autoref{fig:projections}).
We then overlapped each pair of circles $(A,B)$ by adjusting $\angle C_AOC_B$ from $30\degree$ to $0\degree$ with an interval of $1.25\degree$ (the second row of \autoref{fig:projections}). 
Note that unlike Experiment A, the initial projection is the least faithful projection but becomes more faithful as the circles in each pair overlap more.
We used the same metrics setting as Experiment A.

\parahead{Experiment C: Capturing quality degradation} 
To test our metrics' ability to capture the quality degradation of the projection, we computed our and previous metrics for the projections with different levels of quality degradation. 
We first created a  2D $t$-SNE projection of the MNIST dataset \cite{lecun1998mnist} (the first projection on the third row of \autoref{fig:projections}).
We then replaced a certain proportion of the projected points with random points.
We varied the replacement rate from 0 to 100\% with an interval of 5\% (the third row of \autoref{fig:projections}).
The inter-cluster reliability of the projections certainly degrade as the replacement rate increases.
We checked whether the metrics can capture such quality degradation. 
We used the same metrics setting as Experiment A.

\parahead{Experiment D: Identifying the effect of projection hyperparameters}
The final experiment was conducted to evaluate the capability of our metrics to capture the inter-cluster reliability differences caused by the hyperparameter choices of an MDP technique. This experiment was inspired by an analysis from the UMAP paper~\cite{mcinnes2018umap} where the authors assessed the impact of a hyperparameter, the number of \texttt{nearest neighbors} $n$, on the projection quality. Lower $n$ values drive UMAP to more local structures, while higher values make the projection to preserve the global structures rather than the local details. 
In the original analysis, the authors qualitatively analyzed how $n$ affects the UMAP projection of a randomly sampled 3-dimensional RGB cube data. The authors concluded that since randomly sampled data have no manifold structure, larger $n$ values generate more appropriate projections than lower $n$ values. Lower $n$ values instead treat the noises from random sampling as fine-scale local manifold structures, generating an unreliable interpretation of the structure \cite{mcinnes2018umap}.

We tested whether our and previous metrics can quantitatively reproduce this conclusion.
We first constructed a dataset of 4000 points randomly sampled from a 3-dimensional RGB cube. 
UMAP projections of the dataset with different $n$ ($4-9$ with an interval of 1, $10-90$ with an interval of 10) were then generated (the fourth row of \autoref{fig:projections}) and tested with the same metrics setting as Experiment A. 
We set another hyperparameter of UMAP, \texttt{min\_dist}, to 0 because higher \texttt{min\_dist} values tune projections to lose the local structure, reducing the difference between the projections generated with higher and lower $n$ values. Setting it at 0.0 prevents such an effect from affecting the experiment.

\begin{figure}[t]
    \centering
    \includegraphics[width=0.99\linewidth]{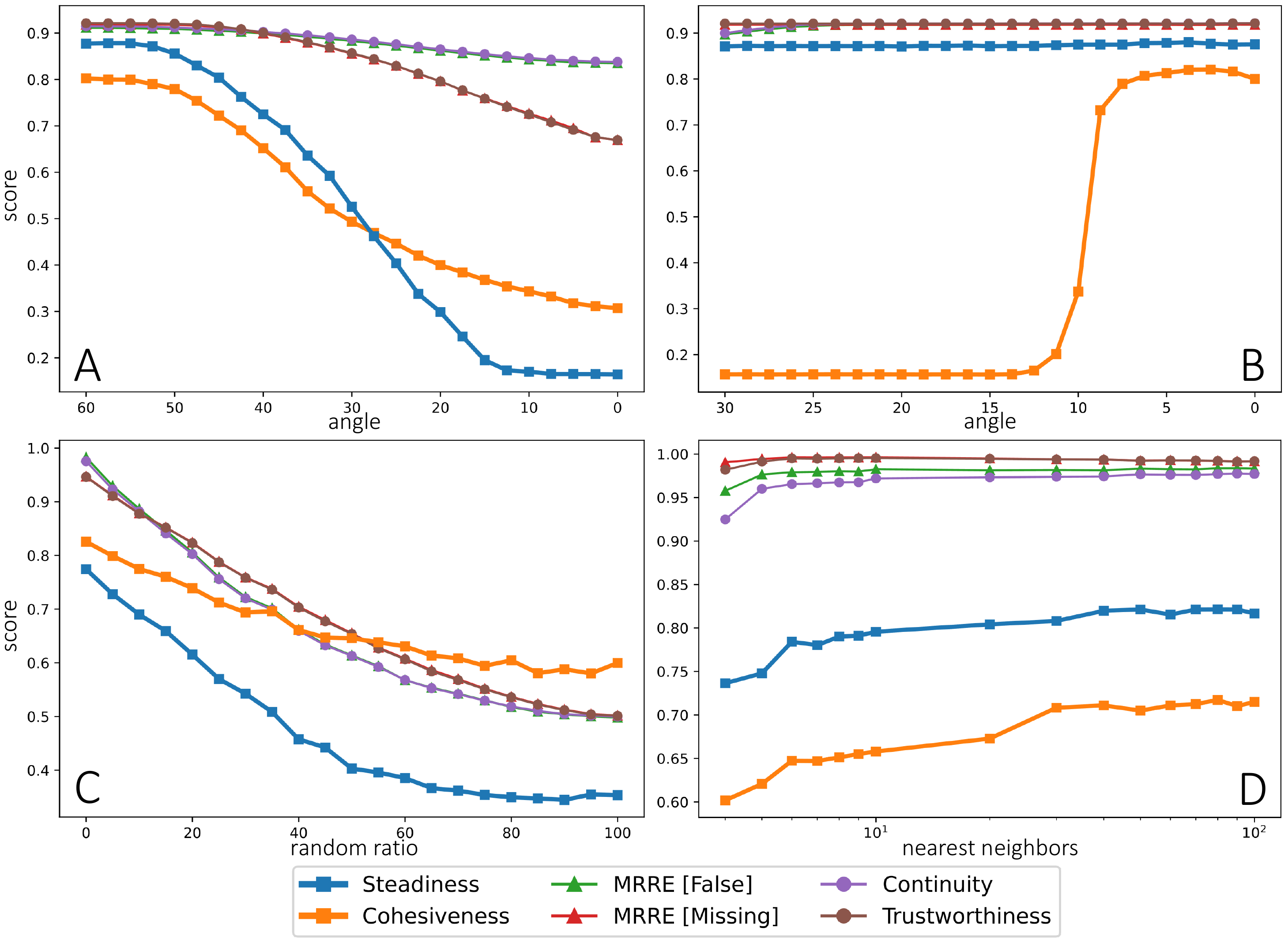}
    \caption{The result of quantitative experiments A-D; the scores measured by our metrics (\snc) and baseline local distortion metrics (MRREs, T\&C).}

    \label{fig:graphs}
\end{figure}

\subsubsection{Results}

\parahead{Experiment A} As we decreased the angle between each circle pair (i.e., increasing the amount of false overlap), both \St (slope $= 3.05 \cdot 10^{-2}$, $p < .001$) and \Co (slope $= 1.98 \cdot 10^{-2}$, $p < .001$) decreased. The baseline local metrics: Trustworthiness (slope $= 9.31 \cdot 10^{-3}$, $p < .001$), Continuity (slope $= 3.02 \cdot 10^{-3}$, $p < .001$), MRRE [Missing] (slope $= 9.13 \cdot 10^{-3}$, $p < .001$), and MRRE [False] (slope $= 3.01 \cdot 10^{-3}$, $p < .001$), also decreased, but the slope was statistically gentle compared the our metrics ($p < .001$ for all).
(\autoref{fig:graphs}A)

\parahead{Experiment B} As we decreased the angles between each circle pair (i.e., making projections more faithful), \Co drastically increased around $10\degree$ (slope $= - 0.169$ in range $[15, 5]$, $p < .001$), which is the point where the circle pair starts to overlap. Other measures such as \St (slope $=-2,29 \cdot 10^{-4}$, $p < .001$), Trustworthiness (slope $= -1.81  \cdot 10^{-5}$, $p < .001$), Continuity (slope $= -7.17  \cdot 10^{-4}$, $p < .001$), MRRE [Missing] (slope $= -3.86  \cdot 10^{-6}$, $p < .001$) and MRRE [False] (slope $= -7.22  \cdot 10^{-4}$, $p < .001$) did not changed significantly (\autoref{fig:graphs}B).

\parahead{Experiment C} As the replacement rate increased, 
\St (slope $= -4.29  \cdot 10^{-3}$, $p < .001$), 
\Co (slope $= -2.35  \cdot 10^{-3}$, $p < .001$), 
Trustworthiness (slope $= -4.60  \cdot 10^{-3}$, $p < .001$), 
Continuity (slope $= -4.70  \cdot 10^{-4}$, $p < .001$), 
MRRE [Missing] (slope $= -4.60  \cdot 10^{-3}$, $p < .001$), and 
MRRE [False] (slope $= -4.78  \cdot 10^{-4}$, $p < .001$) 
all decreased. (\autoref{fig:graphs}C)

\parahead{Experiment D} As we increased $n$, both \St (slope $= 5.78 \cdot 10^{-4}$, $ p < .001$) and \Co (slope $= 9.44 \cdot 10^{-4}$, $ p < .001$) increased, while Trustworthiness (slope $=- 8.17 * 10^{-6}$, $ p < .001$), MRRE [Missing] (slope $=-3.87 \cdot 10^{-5}$, $ p < .001$) decreased. Continuity (slope $=2.13 \cdot 10^{-4}$, $ p < .001$) and MRRE [False] (slope $=8.83 \cdot 10^{-5}$, $ p < .001$) increased, though the slopes were statistically gentle compared to \St ($p < .001$ for both). All baseline local metrics early saturated near the max score around $n=5$. 
(\autoref{fig:graphs}D).

\begin{figure}[t]
    \centering
    \includegraphics[width=\linewidth]{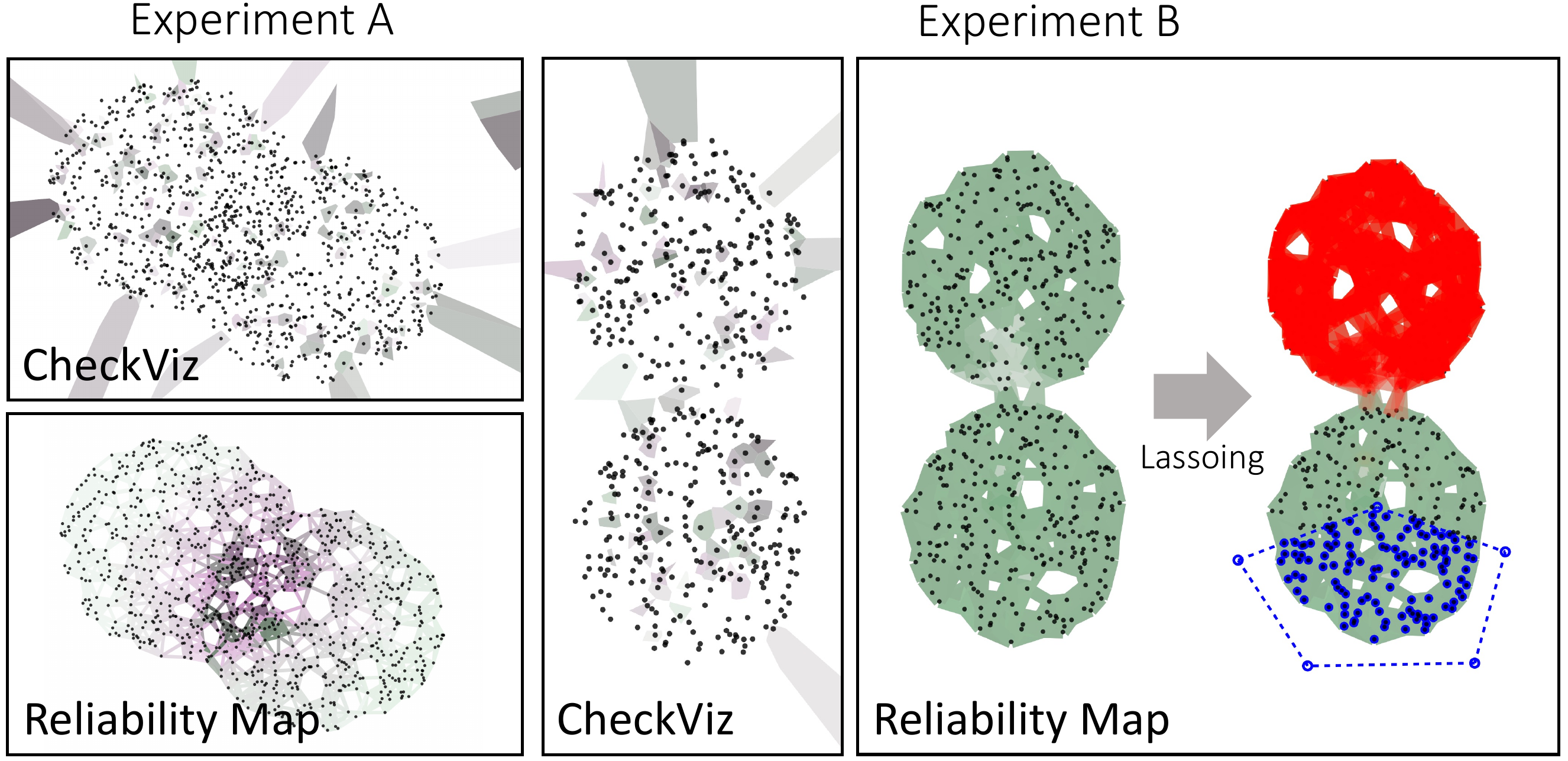}
    \caption{
    The reliability map and CheckViz visualizing the distortion of the projections from Experiment A (red dotted square in \autoref{fig:projections}) and Experiment B (blue dashed square in \autoref{fig:projections})). Unlike CheckViz, where no interesting pattern was shown, the reliability map demonstrated where and how \Fg distortion occurred (Exp. A). Even \Mg distortions was identified by the cluster selection interaction (Exp. B). }

    \label{fig:expexp}
\end{figure}

\begin{figure*}[ht]
    \centering
    \includegraphics[width=\textwidth]{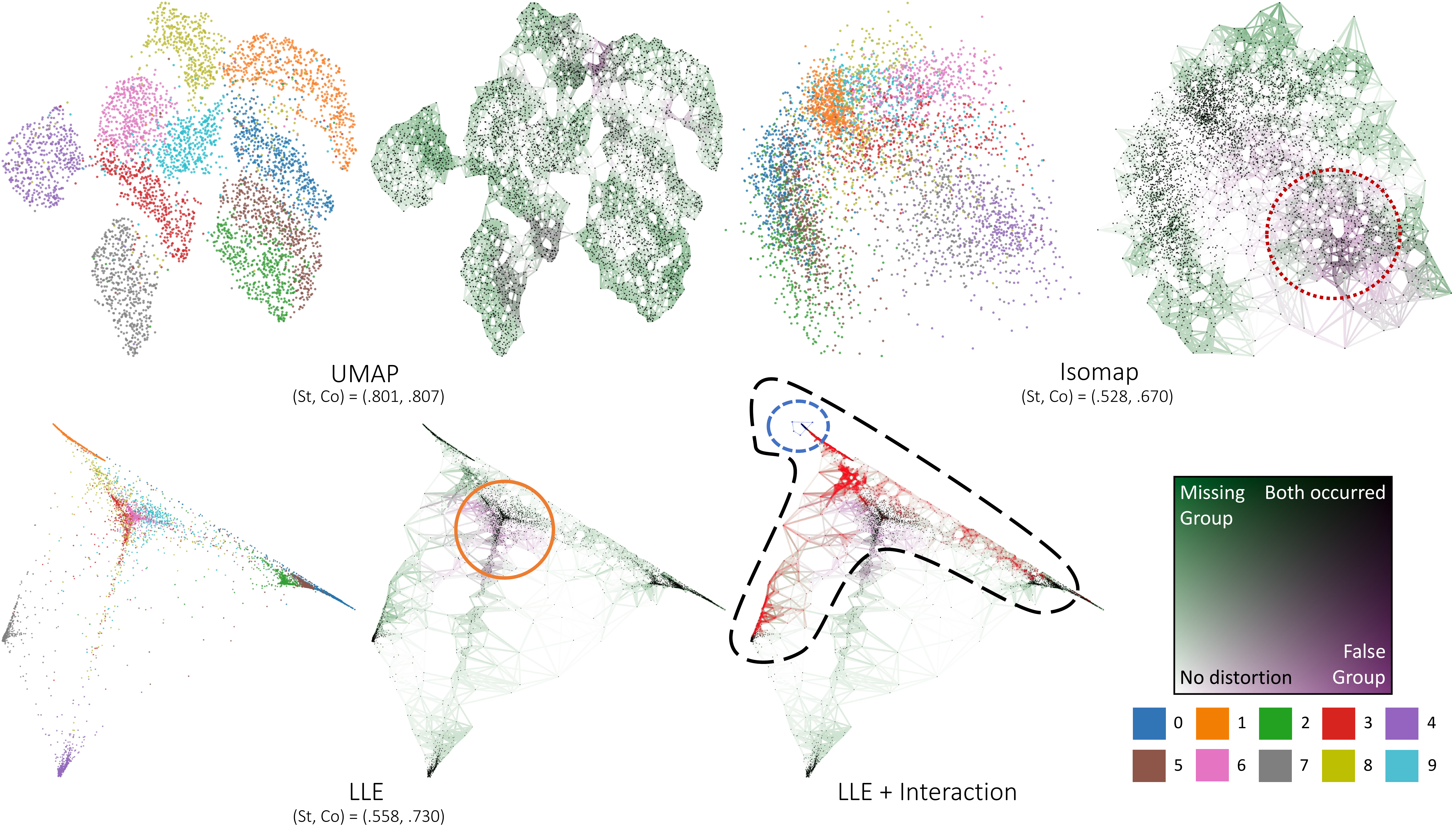}
    \caption{The UMAP, Isomap, and LLE projections of MNIST test dataset and the reliability maps that visualize each projection's inter-cluster distortion. Overall \St (St) and \Co (Co) scores are depicted under the name of each technique. For each MDP technique, the left pane shows the class identity, and the right pane shows the reliability map.  Color-encoding of the inter-cluster distortion used in the reliability map is in the lower right corner. The projection and the corresponding reliability map of $t$-SNE and PCA are in Appendix B.
    }
    \vspace{-3mm}
    \label{fig:mnists}
\end{figure*}

\begin{figure*}[ht]
    \centering
    \includegraphics[width=\textwidth]{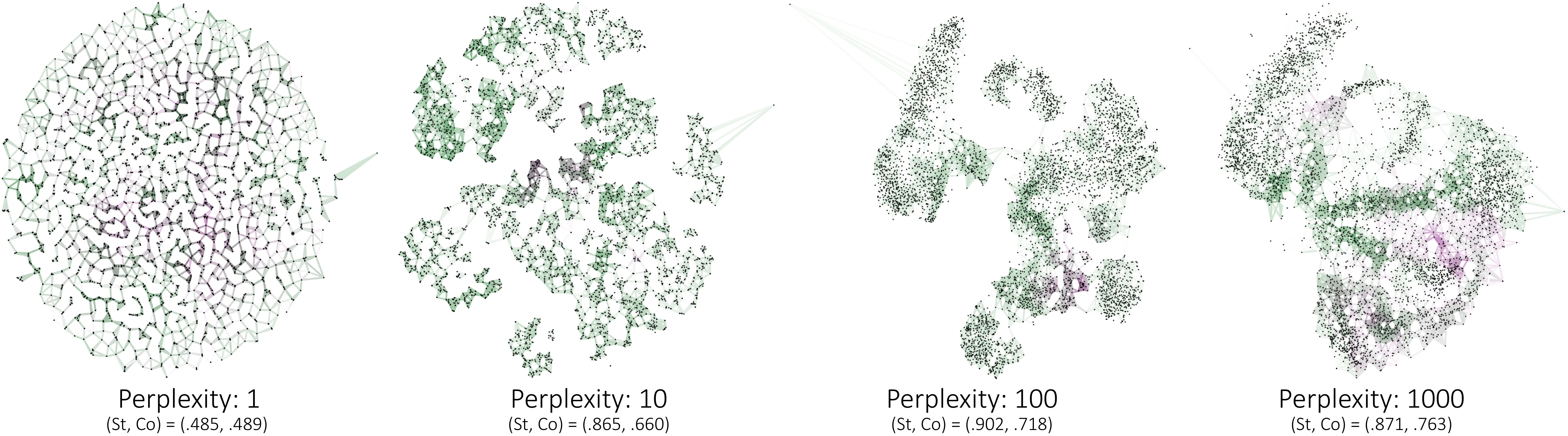}
    \caption{The reliability maps that visualize the inter-cluster distortion of $t$-SNE projections $(\sigma = [1, 10, 100, 1,000])$ made for Fashion-MNIST test dataset. \snc scores are depicted under the the perplexity value of each projection.
    }

    \label{fig:fmnists}
    \vspace{-5.2mm}
\end{figure*}

\subsubsection{Discussion}

The result of Experiment A suggests that our metrics could identify a loss of the inter-cluster reliability caused by \Fg distortion, as \St decreased when the overlap of circle pairs increased. 
\Co also decreased, which means that not only False but also Missing Groups distortions had occurred. This is because for point $p$ and $q$ in a circle, although their Euclidean distance is maintained while the circle is overlapping with another circle, the SNN similarity decreases as more points intervene between $p$ and $q$. Continuity and MRRE [False] also captured the decrease in the inter-cluster reliability due to \Fg distortion, but slower compared to our metrics.  

For Experiment B, the result confirms that \Co correctly identifies \Mg distortion as the metric increased following the increasing overlap of circles that reduces \Mg distortion. Moreover, in Experiment B, the amount of \Mg distortion was captured only by \Co, which showed that our metrics have the ability to pinpoint the particular inter-cluster distortion type.
In contrast, both Trustworthiness and MRRE [Missing] failed to capture this apparent \Mg distortion.

The Reliability map further confirms the results of Experiment A and B as it showed that \snc accurately identified the place where \Fg and \Mg occurred (\autoref{fig:expexp}). Reliability map located the \Fg distortion of Experiment A by highlighting the overlapped area in purple. For Experiment B, it was able to identify the \Mg relationship of two separated circles in a pair through the cluster selection interaction, as selecting the portion of one circle showed that the other circle was actually close to it; this result matches the ground truth. In contrast, CheckViz, which visualized the False and Missing Neighbors distortion of each point computed by T\&C, did not show any pattern.

In Experiments C and D, both \snc could capture the decrease (Experiment C) and the increase (Experiment D) in inter-cluster reliability. Moreover, Experiment D showed that our metrics also can be used to quantify the effect of a hyperparameter by reproducing the result of human observers' qualitative analysis \cite{mcinnes2018umap}. 
In contrast, local metrics barely captured the certain increase of inter-cluster reliability in Experiment D.

Overall, the experiments proved that our metrics can properly measure inter-cluster reliability. On the contrary, local metrics failed for some cases even with the apparent inter-cluster distortion.

\newcommand\Tstrut{\rule{0pt}{2.3ex}}         
\newcommand\TstrutBig{\rule{0pt}{2.8ex}}         
\newcommand\Bstrut{\rule[-1.4ex]{0pt}{0pt}}   

\begin{table}[t]

\centering{
 \begin{tabular}{ccccc}
 \Xhline{1.7\arrayrulewidth} 
 & \multicolumn{4}{c}{Distance Measurement} \Tstrut\\
 \texttt{clustering} & \multicolumn{2}{c}{SNN-based} & \multicolumn{2}{c}{Euclidean} \\
 \cline{2-3} \cline{4-5} 
 & St & Co & St & Co \Tstrut\\
 \hline

HDBSCAN & \small{$5.78 \cdot 10^{-4}$}  &  \small{$9.44 \cdot 10^{-4}$}& \small{$3.35 \cdot 10^{-4}$}&  \small{$-1.37 \cdot 10^{-3}$}  \TstrutBig\\

X-Means & \small{$1.59 \cdot 10^{-4}$}&  \small{$1.10 \cdot 10^{-3}$}&  \small{$2.07 \cdot 10^{-4}$} &   \small{$-4.98 \cdot 10^{-4}$} \TstrutBig\\

 $K$-Means \small{$(K=5)$} & \small{$2.06 \cdot 10^{-4}$} & \small{$1.40 \cdot 10^{-3}$} &  \small{$3.10 \cdot 10^{-4}$}&  \small{$-7.01 \cdot 10^{-4}$} \TstrutBig\\

 $K$-Means \small{$(K=10)$} & \small{$2.36 \cdot 10^{-4}$} & \small{$1.16 \cdot 10^{-3}$} &  \small{$3.36 \cdot 10^{-4}$}&  \small{$-7.29 \cdot 10^{-4}$} \TstrutBig\\

 $K$-Means \small{$(K=20)$} & \small{$2.91 \cdot 10^{-4}$} & \small{$9.32 \cdot 10^{-4}$} &  \small{$3.41 \cdot 10^{-4}$}&  \small{$-6.12 \cdot 10^{-4}$} \TstrutBig\\

 \Xhline{1.7\arrayrulewidth}
\end{tabular}
\vspace{3pt} 
\caption{\label{tab:robust} The result of Experiment D conducted with diverse hyperparameter procedure settings (\autoref{sec:6rob}). Each cell depicts the slope of regression line which represents the relation between \texttt{nearest neighbor} value and the score of \St (St) and \Co (Co). Every regression analysis result satisfied $p < .001$.}}

\vspace{-0.5mm}
\end{table}

\subsection{Robustness Analysis}

\label{sec:6rob}

We also investigated the robustness of \snc against hyperparameters by conducting Experiment D using \snc with different hyperparameters. We tested \snc with simpler hyperparameters functions as hyperparameter functions can considerably change the behavior of our metrics.
For the goal, we tested simpler clustering algorithms, X-Means \cite{pelleg2000x} and $K$-Means \cite{duda1973pattern} (number of clusters $ = 5, 10, 20$), instead of the default HDBSCAN algorithm for \texttt{clustering}. We also tested the Euclidean distance as \texttt{dist} instead of the default SNN-based distance.
While using Euclidean distance for the distance measurement between points, we also defined the distance between two clusters \texttt{dist\_cluster} as the Euclidean distance between their centroids instead of the default definition based on SNN similarity to align with \texttt{dist}. For \texttt{extract\_cluster}, we treated every traversed points as cluster members instead of using probability to weight the points with high SNN similarity.

As a result (\autoref{tab:robust}), \snc with simpler \texttt{clustering} hyperparameter functions both increased as \texttt{nearest neighbors} values increased, which confirms the ability to properly quantify inter-cluster reliability. This result shows that our metrics' capability is not bound mainly by the selection of \texttt{clustering} but instead originates more from the power of randomness to analyze complex structures \cite{tempo2012randomized}. 
What is interesting here is that the case of \textit{K}-Means ($K=20$) showed the most similar results to the case of the default HDBSCAN hyperparameter both for \snc. This is because when clustering the extracted clusters in the opposite space, the inter-cluster structure composed of arbitrary shapes and sizes can be better represented by the fine-grained \textit{K}-Means clustering result than the coarse-grained result. 

However, \Co failed for all cases that used Euclidean distance as \texttt{dist}. This result shows that while designing hyperparameters functions for \snc, users should carefully consider the definition of the distance between two points.

\section{Case Studies}

\label{sec:7777}

We report two case studies that we conducted with two ML engineers (E2, E3). During the study, we demonstrated to the engineers how \St, \Co, and the reliability map works, and they explored with us the original inter-cluster structure of MNIST and Fashion-MNIST \cite{xiao2017fashion} test datasets, where both live in a 784-dimensional space and consist of 10 classes. 
The case study showed that our metrics and the reliability map supports users in 1) selecting adequate projection techniques or hyperparameter settings that match the dataset and 2) preventing users' misinterpretation that could potentially occur while conducting inter-cluster tasks (\autoref{sec:3}). 
ML engineers agreed that such support is helpful in interpreting the inter-cluster structure of high-dimensional data.

\subsection{MNIST Exploration with Diverse MDP Techniques} 

\label{sec:71}

To explore the inter-cluster structure of MNIST, we projected it with  
$t$-SNE, UMAP, PCA \cite{pearson1901liii}, Isomap \cite{tenenbaum2000global}, and LLE. 
We measured the \snc ($k=100$, 500 iterations) of each projection and visualized the result with the reliability map (\autoref{fig:mnists}). 

We first discovered that visualizing \snc can prevent users from misidentifying a cluster separation of the original space (T1). For instance, in the Isomap projection, we found the region with high \Fg distortion consists of categories \#4 and \#7 (red dotted circle in \autoref{fig:mnists}). A similar region was also observed in the PCA projection (Appendix B). 
LLE also has the cluster with high \Fg distortion composed of categories \#3, \#6, \#8, and \#9.
Without checking \Fg distortion, one could make the wrong interpretation that such a region belongs to same cluster;
 visualizing the distortion with the reliability map helped us avoid this misperception.  

Visualizing \Fg distortion also allowed additional reasoning beyond a mere quantitative score comparison to choose the proper projection technique. We found that \Fg distortions that occurred in Isomap, PCA (overlap of category \#4 and \#7), and LLE (overlap of category \#3, \#6, \#8, \#9) did not occur in $t$-SNE or UMAP.
This finding explains why the \St of $t$-SNE and UMAP are higher than other projections, advocating the use of $t$-SNE and UMAP in exploring the inter-cluster structure of a MNIST dataset.

Still, as the $t$-SNE and UMAP projections also suffered from \Mg distortion, we critically interpreted that the clusters in these projections actually stay closer to each other than they look. E3 noted that this interpretation matches the ground truth that digits in MNIST stay much closer and mixed in the original space than their representations.

Moreover, we found that by using cluster selection interaction, users can accurately estimate and compare cluster sizes and shapes (T3). As we selected the local area in LLE (blue dashed ellipse in \autoref{fig:mnists}), the reliability map highlighted a much larger region around the selected region (black long-dashed contour in \autoref{fig:mnists}). This means that the original cluster containing the selected points was much larger than we can see in the projections and lost its portion as dispersed outliers (i.e., \Mg distortion occurred). We identified this problem through cluster selection interaction and escaped from the misinterpretation.

\subsection{Fashion-MNIST Exploration with \textit{t}-SNE}
 
In the second case study, we explored Fashion-MNIST dataset using $t$-SNE projections with varying hyperparameters. We measured our metrics ($k=100$, 500 iterations) on $t$-SNE projections generated with different perplexity values $\sigma \in [1, 10, 100, 1,000]$ and visualized the result with the reliability map (\autoref{fig:fmnists}).
Note that using a high value for $\sigma$ makes the $t$-SNE focus more on preserving global structures \cite{wattenberg2016how}.

As a result, we found that our metrics and the reliability map can help in selecting adequate hyperparameter settings. For example, the projection with $\sigma=1$ had both False and \Mg distortion distributed uniformly across the entire projection space. This finding, which aligns with the low score of the projection, showed that low $\sigma$ values are not sufficient to capture the global inter-cluster structure, which matches its actual behavior. This result justifies that it is proper to select a higher $\sigma$ value to investigate Fashion-MNIST. The fact that $\sigma=100$ and $\sigma=1,000$ projections earned higher scores for both \St and \Co compared to $\sigma=1$ and $\sigma=10$ projections strengthens this interpretation.

Thus, we subsequently analyzed the $\sigma=100$ and $\sigma=1,000$ projections and discovered that our metrics prevent users from making the wrong interpretation of the relations between clusters (T2). 
We first noticed that the $\sigma=100$ projection has more compact clusters compared to the projection with $\sigma=1,000$, where clusters are slightly more disperser and closer to each other. As the $\sigma=100$ projection achieved a relatively high \St score, we could conclude that each compact cluster also exists as a cluster also in the original space.
However, as the $\sigma=1,000$ projection increased \Co, we were not able to believe the separation of the clusters depicted in the $\sigma=100$ projection (T2-1). According to \Co, the distances between the clusters in the original spaces is better depicted in the $\sigma=1,000$ projection. Therefore, it is more reliable to interpret the original inter-cluster structure as a set of subclusters that constitute one large cluster rather than as a set of separated clusters (T2-2). E2 paid particular attention to this result. She pointed that it is common to perceive that projections with well-divided clusters (e.g., $\sigma=100$ projection) better reflects the inter-cluster structure, but this result shows that such a common perception could lead to a misinterpretation of inter-cluster structure.

\section{Conclusion and Future Work} 

Although it is important to investigate the inter-cluster distortion in many MDP tasks, there were previously no metrics that directly measure such distortions.
In this work, we first surveyed user tasks related to identifying the inter-cluster structures of MDP and elicited design considerations for the metrics to evaluate the inter-cluster reliability.
Next, we presented \snc to evaluate inter-cluster reliability by measuring the \Fg and \Mg distortions and presented the reliability map to visualize the metrics. 
Through quantitative evaluations, we validated that our metrics adequately measure inter-cluster reliability. The qualitative case studies showed that our metrics can also help users select proper projection techniques or hyperparameter settings and perform inter-cluster tasks with fewer misperceptions, assisting them in interpreting the original space's inter-cluster structure.

As a future work, we plan to enhance the scalability of our algorithm. The algorithm currently computes the iterative partial distortion measurement sequentially.
As each iteration works independently, we plan to accelerate the algorithm leveraging multiprocessing. 
We also plan to improve our metrics to consider the hierarchical aspect of inter-cluster structures and reduce the number of hyperparameters.
Another interesting research direction would be to investigate how \St, \Co, and their visualizations affect users' perception of the original data, which will provide an in-depth understanding of MDP, as an expansion of our case study.

\acknowledgments{
Thanks to Yoo-Min Jung and Aeri Cho for their valuable feedback.
This work was supported by the National Research Foundation of
Korea (NRF) grant funded by the Korea government (MSIT) (No.
NRF-2019R1A2C208906213).}

\bibliographystyle{abbrv-doi}

\typeout{}

\bibliography{main}
\end{document}